\documentclass{article} 
\usepackage[final]{colm2026_conference}

\usepackage{microtype}
\usepackage{hyperref}
\usepackage{url}
\usepackage{booktabs}

\usepackage{booktabs}   
\usepackage{multirow}   
\usepackage{graphicx}   
\usepackage{colortbl}   

\usepackage{subcaption}
\usepackage{caption}
\usepackage{xcolor}     

\usepackage{booktabs}
\usepackage{tabularx}
\usepackage{array}

\usepackage{graphicx}
\usepackage{subcaption}

\usepackage[most]{tcolorbox}
\usepackage{xcolor}
\usepackage{listings}
\usepackage{caption}
\usepackage{xltabular}

\definecolor{codebg}{HTML}{202020}
\definecolor{codetext}{HTML}{F2F2F2}
\definecolor{codeframe}{HTML}{111111}

\newtcblisting{verifierbox}{
  listing only,
  enhanced,
  colback=codebg,
  colframe=codeframe,
  coltext=codetext,
  colupper=codetext,
  arc=9pt,
  boxrule=0.8pt,
  left=10pt,
  right=10pt,
  top=10pt,
  bottom=10pt,
  before skip=6pt,
  after skip=6pt,
  listing options={
    basicstyle=\ttfamily\scriptsize\color{codetext},
    columns=fullflexible,
    breaklines=true,
    keepspaces=true,
    showstringspaces=false,
    frame=none
  }
}

\usepackage{lineno}

\definecolor{darkblue}{rgb}{0, 0, 0.5}
\hypersetup{colorlinks=true, citecolor=darkblue, linkcolor=darkblue, urlcolor=darkblue}

\title{BudgetVerify: Budget-Tiered Verification for Financial QA}

\author{Janet Jenq and Hongda Shen \\
PitchBook\\
\texttt{\{janet.jenq,hongda.shen\}@pitchbook.com} \\
}

\begin{document}

\ifcolmsubmission
\linenumbers
\fi

\maketitle

\begin{abstract}
Financial question answering often requires precise numerical extraction, unit handling,
and arithmetic over tables and text, but applying expensive verification uniformly wastes
test-time compute. We propose BudgetVerify, a budget-tiered generator--verifier framework
that routes each generated answer to one of three verification tiers: no verification, lightweight
check-and-revise, or higher-cost solve-first-then-compare verification. The router is trained from offline correctness and token-cost outcomes and, at test time, selects a verification tier using information available before verification, including the question, context statistics, the generated answer, and associated generator metadata. The selected tier either returns the generated answer directly or invokes the corresponding verifier. Across six commercial and open-weight base models, BudgetVerify consistently produces more efficient accuracy-cost Pareto frontiers than fixed verification policies by selectively allocating stronger verification only when it is useful. Although absolute performance varies across models, these efficiency gains and the resulting qualitative frontier shape are consistent across generator models.
\end{abstract}

\section{Introduction}
\label{sec:intro}

Financial question answering (QA) is a critical capability for analyst-facing LLM systems, with applications ranging from earnings analysis and financial statement review to risk assessment, due diligence, and automated report generation. Unlike open-domain QA, financial QA often requires the precise extraction of values from tables and text, the careful handling of units and scales, and aggregation arithmetic over specific figures such as margins, growth rates, and ratios \citep{chen-etal-2021-finqa, islam2023financebenchnewbenchmarkfinancial, Jenq_2026_CVPR}. This makes answer accuracy measurement highly sensitive to small errors such as selecting the wrong period, using the wrong denominator, or confusing percentages with ratios, since it can change the final answer even when the model’s response appears well-reasoned. At the same time, financial questions vary widely in difficulty, ranging from a direct table lookup to requiring multi-step numerical reasoning, suggesting that a single fixed inference strategy may be inefficient or unsuccessful.

\begin{figure}[h!] 
 \center{\includegraphics[width=\textwidth]{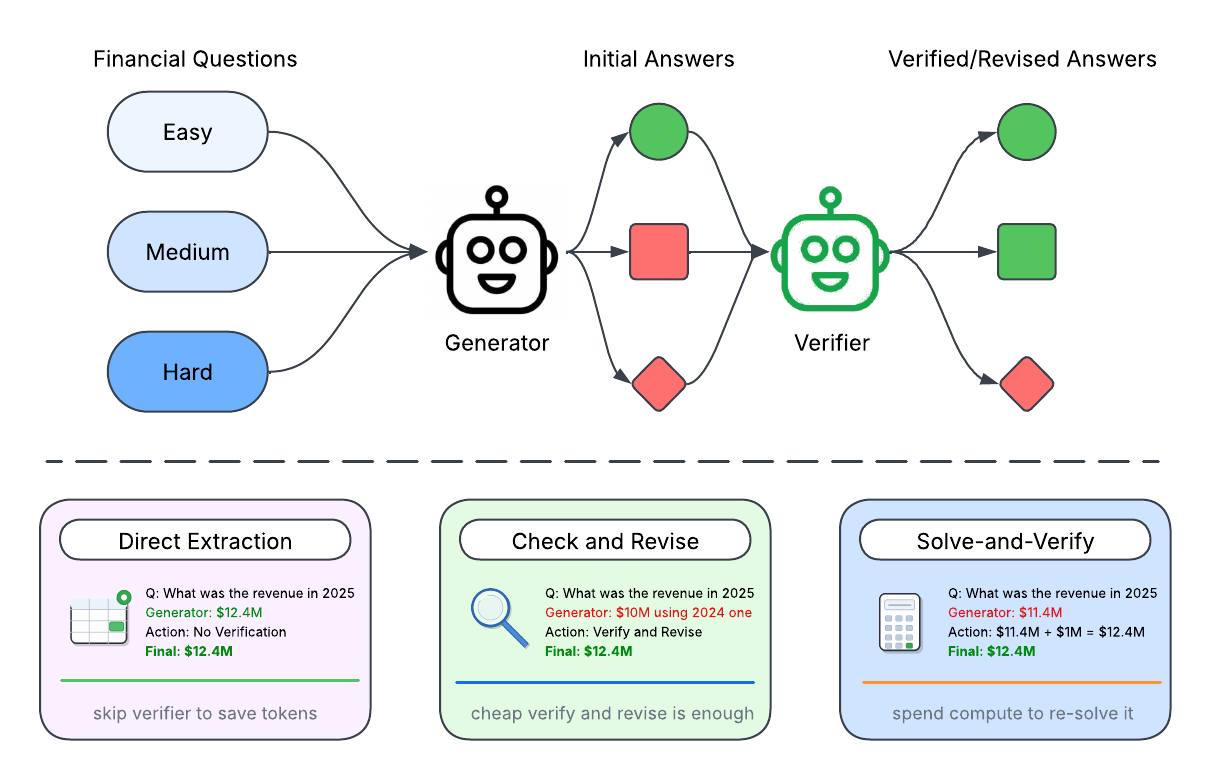}}
 \caption{\textbf{Generator–verifier workflow and FinQA verification conditions.} The top panel shows the standard generator–verifier paradigm, where generated answers are checked or revised by a verifier. Green markers denote correct answers, while red markers denote incorrect answers. The bottom panels illustrate three conditions commonly observed in FinQA: direct extraction often needs no verification, simple errors can be fixed by lightweight checking, and harder numerical reasoning may require solve-and-verify computation.}
 \label{fig:intro}
 \end{figure}

Test-time scaling (TTS) offers a way to improve answer quality by allocating additional compute resources during inference rather than changing model parameters. A common instantiation is the generator-verifier paradigm \citep{pandit2025hard2verifysteplevelverificationbenchmark}, in which a generator first produces one or more candidate answers and a verifier then evaluates, filters, or refines these candidates before determining the final answer. Figure~\ref{fig:intro} illustrates this: generated answers may be correct or incorrect, and verification aims to either preserve correct candidates or fix wrong ones. However, generator-verifier systems introduce a compute allocation problem: one can spend budget on generating more candidates to apply stronger verification, or repeatedly refining an answer, but these decisions increase cost and latency. Standard test-time pipelines often apply a uniform strategy to all inputs, such as generating the same number of candidates or invoking the same verifier regardless of query difficulty. Recent work on verification dynamics \citep{zhou2026variationverificationunderstandingverification} shows that verification success is not determined solely by the quality of the verifier. It depends jointly on the difficulty of the problem, the capability of the generator, and the strength of the verifier's problem-solving ability. In particular, verifiers are more likely to recognize correct responses on easier problems, errors from weaker generators are easier to detect than errors from stronger generators, and improving verifier quality may see limited gains in certain settings. These findings suggest that always applying the strongest verifier can misallocate inference compute, where some generated answers may be cheaply verified, while others may remain difficult even under stronger verification. This motivates our central question: \textit{given a generated financial answer, how much verification compute should be spent?}

This allocation problem is especially relevant in financial question answering, where inputs vary in the amount and type of computation they require. As shown in the bottom panels of Figure~\ref{fig:intro}, some questions can be answered from a single extracted value and may not need verification, while others contain local generator errors that can be corrected through a simple check. More challenging cases require combining information across tables, text, periods, or numerical operations, thereby justifying a higher-cost verifier. As a result, a fixed verification strategy is unlikely to be optimal. Lightweight checking may be sufficient for some generated answers, whereas others may require a more expensive verifier that reconstructs the answer more carefully. We therefore formulate financial QA verification as a budgeted routing problem. Given a generated answer, BudgetVerify adaptively selects among multiple verification tiers, ranging from no verification to lightweight revision to solve-first-then-compare verification, with the goal of improving aggregate accuracy under a fixed inference budget.

We evaluate this framework across multiple base models, not to compare or rank their absolute capabilities, but to test whether the same accuracy–cost structure recurs across different model settings. The central empirical finding is that it does. Although the models differ in baseline accuracy and token cost, every evaluated model exhibits the same qualitative Pareto-frontier shape: selective verification produces substantial early accuracy gains, followed by diminishing returns as increasingly expensive verification is applied. Thus, the absolute position of the frontier is model-dependent, while its qualitative shape is robust across base models.

\section{Related Work}
\label{sec:related_work}

\textbf{Test-Time Scaling.} Test-time scaling improves model performance by spending additional computation at inference time, for example through repeated sampling, revision, search, or verifier-guided selection, rather than changing model parameters \citep{wang2023selfconsistencyimproveschainthought, snell2024scalingllmtesttimecompute,brown2024largelanguagemonkeysscaling,gong-etal-2025-mice}. Recent work \citet{zhou2025evaluatingjudgesevaluatorsjetts,zhai2026adaptivetesttimecomputeallocation} shows that the value of additional inference compute is highly non-uniform across inputs: compute-optimal allocation can outperform uniform best-of-N strategies and, in some cases, provide larger gains than simply scaling model size. \citet{snell2024scalingllmtesttimecompute} study how to allocate test-time compute across prompts and show that adaptive compute allocation can substantially improve efficiency over fixed-budget baselines, while more recent work formulates adaptive test-time compute allocation as a constrained optimization problem and trains lightweight policies to imitate oracle budget decisions. Our work follows this view but studies compute allocation across a different axis: rather than deciding how many samples to draw, we decide how much verification compute to spend on a generated financial answer.

\textbf{Generator - Verifier Paradigm.} A common mechanism for using test-time compute is the generator–verifier framework, where a generator proposes candidate solutions and a verifier scores, filters, or refines them \citep{lightman2023letsverifystepstep, cemri2025multiagentllmsystemsfail}. This paradigm appears in best-of-N selection \citep{zhang2025generativeverifiersrewardmodeling,snell2024scalingllmtesttimecompute}, process- and outcome-supervised verification, and generative verifiers that produce verification rationales before judging candidate answers. Prior work shows that step-level/process supervision can improve multi-step reasoning, while generative verifiers can outperform discriminative verifiers and LLM-as-judge baselines in reasoning settings \citep{zha2025rltangoreinforcinggenerator}. Recent analysis by \citet{zhou2026variationverificationunderstandingverification} further shows that verification effectiveness depends on problem difficulty, generator capability, and verifier capability, suggesting that verification should not be treated as a uniform post-processing step. In contrast to work that primarily scales the number of candidates or verifier votes, we study budgeted verification depth for a single generated financial answer, ranging from no verification to lightweight revision to solve-first verification \citep{sun2025s2jbridginggapsolving, chen2026rmr1rewardmodelingreasoning}.

\textbf{Routing and Cost-Aware Allocation.} Routing has recently emerged as an effective way to improve LLM systems without updating the underlying model weights. Systems such as vLLM Semantic Router~\citep{vllm_semantic_router} route requests using signals from the request, response, and context to improve efficiency and safety, while Sakana Fugu~\citep{sakanai_fugu} exposes a single model-like interface that dynamically orchestrates specialized LLM agents for each query. In a related line of work, routers choose among models, tools, or compute budgets to balance quality and cost. FrugalGPT \citep{chen2023frugalgptuselargelanguage} introduced methods that route queries through cheaper or stronger LLM layers to reduce inference cost, RouteLLM \citep{ong2025routellmlearningroutellms} learns routers from preference data to dynamically choose between weaker and stronger LLMs, and CARROT \citep{somerstep2025carrotcostawarerate} frames routing as a prediction problem where each option's cost and performance are estimated before proceeding with the option with the best perceived tradeoff. These works motivate our supervised routing formulation: we first run multiple verification tiers offline to obtain correctness and cost outcomes, derive cost-aware oracle labels, and then train a router model to choose among verification options at inference time. Unlike model-routing methods in prior work, which select a base LLM for each query, BudgetVerify fixes the base model and allocates verification compute after an initial answer has been generated. The routing decision therefore focuses on verification depth rather than model choice. This distinction is relevant in financial QA because verification needs depend not only on question difficulty, but also on the structure of the generated error. A stronger model may lower the average error rate, but may not indicate whether a given answer should be retained, locally revised, or entirely re-solved. For example, a copied value or arithmetic inconsistency may be corrected with a lightweight check, whereas a period mismatch, missing operand, or denominator error may require re-solving the problem from the source context. BudgetVerify aims to determine the appropriate verification action for each question and generated answer pair.

\section{Methodology}
\label{sec:methodology}

\subsection{Problem Formulation}

We study financial question answering in a single-candidate generator-verifier setting. Given a question $q$ and financial context $c$, a generator $G$ first produces an initial structured answer $a_0 = G(q,c)$. Rather than applying verification uniformly, the system must choose a verification tier $t$ from a finite set $\mathcal{T} = \{\textsc{Low}, \textsc{Mid}, \textsc{High}\}$, where \textsc{Low} returns the generator answer directly without verification, \textsc{Mid} applies a lightweight check-and-revise verifier, and \textsc{High} invokes a higher-cost solve-first-then-compare verifier. Each tier produces a final answer $a_t(q,c,a_0)$ with an associated inference cost $C_t(q,c,a_0)$, including the generator token cost. Since the generator is
always run before routing, this generator cost is constant across tiers for a fixed example. We include it in $C_t$ to match the total token cost reported in experiments. Let $\mathrm{Correct}(a_t(q,c,a_0), y) \in \{0,1\}$ denote whether the final answer matches the ground truth execution answer $y$. Our goal is to learn a lightweight routing policy $\pi(q,c,a_0) \in \mathcal{T}$ that selects a verification tier for each generated answer so as to maximize a cost-sensitive utility that trades off final answer accuracy and compute/token cost. We formulate this as a cost-sensitive objective:
\begin{equation}
\max_{\pi} \; \mathbb{E}_{(q,c,y)}
\left[
\mathrm{Correct}\!\left(a_{\pi(q,c,a_0)}, y\right)
-
\lambda C_{\pi(q,c,a_0)}(q,c,a_0)
\right]
\end{equation}

where $\lambda$ controls the accuracy-cost tradeoff. This formulation treats verification as an adaptive test-time resource: the router should avoid unnecessary verification for easy answers, while allocating stronger verification to generated answers whose expected improvement justifies the additional cost.

\begin{figure}[h!] 
 \center{\includegraphics[width=\textwidth]{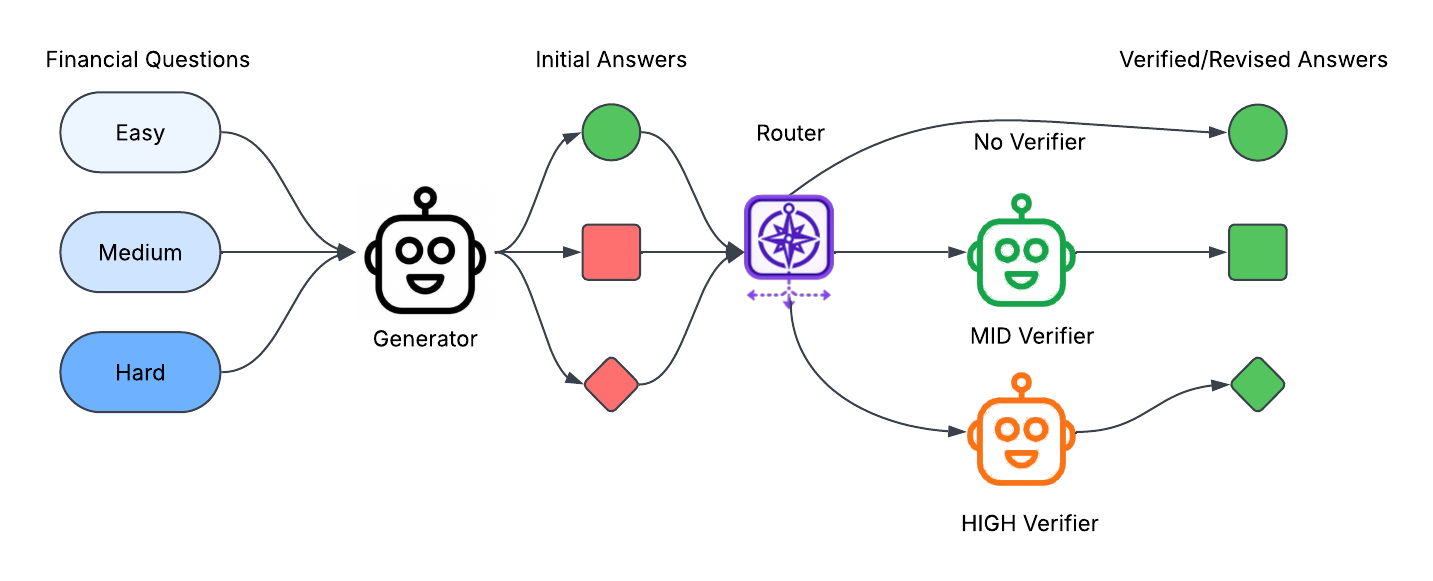}}
 \caption{\textbf{Overview of BudgetVerify.} A generator first produces an initial answer for each financial question. A lightweight router then uses pre-verification features from the question and generated answer to choose among three tiers: no verification, a low-cost MID verifier, or a higher-cost HIGH verifier. Green markers denote correct answers and red markers denote incorrect answers. The router allocates verification compute adaptively, aiming to preserve easy correct answers while sending harder or error-prone cases to stronger verification.}
 \label{fig:budgetverify}
 \end{figure}

\subsection{BudgetVerify}

BudgetVerify consists of a generator, three verification tiers, and a lightweight supervised
router. Figure~\ref{fig:budgetverify} illustrates the overall workflow. For each input
$(q,c)$, the generator first produces a structured answer $a_0 = G(q,c)$ containing the
final prediction and metadata such as the predicted operation, answer unit, formula,
operands, evidence snippets, and confidence. The router then selects one tier from $\mathcal{T}$.

To train the router, we construct an offline verification dataset by running all three tiers
on the training split and recording each tier's final answer correctness and token cost.
For each value of the cost parameter $\lambda$, we derive a cost-aware oracle label
\begin{equation}
\label{eq:tier}
t^*_\lambda(q,c,a_0)
=
\arg\max_{t \in \mathcal{T}}
\left[
\mathrm{Correct}(a_t(q,c,a_0),y)
-
\lambda C_t(q,c,a_0)
\right].
\end{equation}


We then train lightweight multiclass classifiers to imitate these oracle tier choices.
Although the oracle labels are derived from offline runs of all three verification tiers,
the resulting verifier outcomes are used only as supervision targets. During both training
and inference, the router's input features are restricted to pre-verification information
available before \textsc{Mid} or \textsc{High} is executed. The router never observes verifier outputs, verifier correctness, oracle labels, or post-verification metadata as inputs. Pre-verification features include question-level cues such as length, number count, period count, percentage
cues, and financial terms; context-light statistics such as table size and numeric density; and generator-output features such as predicted operation, answer unit, formula presence, operand count, evidence coverage, and confidence. This design allows BudgetVerify to allocate verification compute selectively while keeping routing overhead small relative to verifier cost. The full feature dictionary is provided in Appendix Table~\ref{tab:router_features}.

\section{Experimental Results}
\label{sec:experiments}

\subsection{Datasets and Experimental Configuration}
We evaluate BudgetVerify on FinQA~\citep{chen-etal-2021-finqa}, a financial question answering benchmark built from company financial reports. Each sample consists of a question, surrounding textual evidence, a financial table, a gold executable answer, and an annotated reasoning program. FinQA is well suited for our experiments because many samples require numerical operations over heterogeneous evidence, including table lookup, arithmetic over extracted values, ratios, percentage changes, and multi-step computations. We use the official train, validation, and test splits throughout, which contain non-overlapping input reports, and evaluate final answers using execution accuracy against the gold executable answer. For each base model, we run the generator and all three verification tiers on the training split to construct the offline routing dataset, including the correctness and token cost. Routers are trained only on the training split. We use the validation split for tuning router hyperparameters. Final accuracy–cost results are reported on the official test split. No samples from the test split are used for router training, feature selection, or hyperparameter tuning.

\begin{table}[ht!]
\centering
\small
\begin{tabular}{l r}
\toprule
\textbf{Statistics} & \textbf{Value} \\
\midrule
Total examples & 8,281 \\
Train/Validation/Test & 6,251/883/1,147 \\
Report pages & 2,789 \\
Avg. input text tokens & 628.11 \\
Avg. table tokens & 59.42 \\
Avg. total input tokens & 687.53 \\
Avg. question length & 16.63 \\
\bottomrule
\end{tabular}
\caption{Basic statistics of the FinQA dataset. FinQA contains financial question-answer pairs with structured tables, unstructured text, executable answers, and annotated reasoning programs.}
\label{tab:finqa_stats}
\end{table}

We evaluate BudgetVerify across two model families: commercial API models and open-weight models. For the commercial setting, we use \texttt{gpt-5.4-nano}, \texttt{gpt-5.4-mini}, and \texttt{gpt-5.4}, which provide a range of cost and capability levels. For the open-weight setting, we use Qwen-family models, including \texttt{Qwen-3.5-4B}, \texttt{Qwen-3.5-9B}, and \texttt{Qwen-3.6-35B-A3B}. To keep the experimental design focused on verifier budget allocation rather than model ensembling, we use a single generator sample per question and restrict the generator and verifier to use the same base model within each run. That is, for a given model, the generator first produces one structured answer, and the \textsc{Mid} and \textsc{High} verification tiers are instantiated using the same model family and checkpoint/API. We provide the complete \textsc{Mid} and \textsc{High} verifier templates in Figures~\ref{fig:mid_template} and~\ref{fig:high_template} in the Appendix. This controlled setup isolates the effect of BudgetVerify's routing decision, whether to skip verification, apply lightweight checking, or invoke higher-cost solve-first verification, without introducing confounding factors from cross-model generator-verifier pairings or multiple candidate sampling.


BudgetVerify is compared against three fixed verification baselines that serve as reference
points on the accuracy--cost tradeoff. \textsc{Generator Only} corresponds to the
\textsc{Low} tier and returns the initial generated answer without verification. Since
verifiers can range from lightweight self-checks to independent re-solving, tool-augmented
verification, or multi-round critique, we instantiate two representative verifier tiers along
a cost--capability axis. \textsc{Always Mid} applies the lightweight check-and-revise verifier
to every example. It is candidate-centered: it defaults to keeping the generator answer and
revises only obvious local errors, such as malformed outputs, copied wrong values, or
single-step arithmetic inconsistencies in the candidate's stated operands. It does not
re-solve the problem, infer missing operands, or search for subtle denominator, sign, scale,
or formula errors. \textsc{Always High} applies the higher-cost solve-first-then-compare verifier uniformly.
It is problem-centered: it first solves the question independently from the context, extracts
a structured operation and operands, executes the computation with a deterministic
calculator, checks operand grounding, and then compares the computed answer with the
generator's candidate. These baselines test whether adaptive routing can approach the
accuracy of stronger fixed verification policies while using fewer tokens on average.

For BudgetVerify, the cost parameter $\lambda$ controls the accuracy--cost tradeoff:
smaller values route more samples to \textsc{Mid} or \textsc{High}, while larger values
favor cheaper \textsc{Low} decisions. For each base model and router class, we sweep a
fixed non-uniform grid of 26 $\lambda$ values in $[0,1]$, selected using only train/validation
runs before test evaluation and made denser near zero. Each value of $\lambda$ defines a
different set of oracle labels via Eq.~\ref{eq:tier}. We train a separate router
$\pi_\lambda$ for each $\lambda$ using only pre-verification features from the training split.
Router hyperparameters are tuned on validation and then fixed. On the official FinQA test
split, each trained router assigns every sample to exactly one tier, and the final answer
and token cost are taken from the selected tier only. We precompute all tier outputs on the
test split only to evaluate fixed baselines and counterfactual routed policies. A deployed
BudgetVerify system would execute only the tier selected by $\pi_\lambda$. Token cost
includes generator input/output tokens and, when \textsc{Mid} or \textsc{High} is selected,
the corresponding verifier input/output tokens. We report normalized token cost by
dividing the routed policy's total token cost by the generator-only total cost on the same
test samples. Accuracy is the fraction of test samples for which the selected tier's final
answer matches the gold executable answer. Figure~\ref{fig:pareto_all_models} plots one
test-set accuracy--cost point for each $\pi_\lambda$ and reports the empirical Pareto frontier
by retaining points for which no other evaluated router achieves at least as high accuracy
at no greater token cost. Fixed baselines are shown as reference points corresponding to
always selecting \textsc{Low}, \textsc{Mid}, or \textsc{High}.

\begin{figure}[h]
\centering
\includegraphics[width=0.99\linewidth]{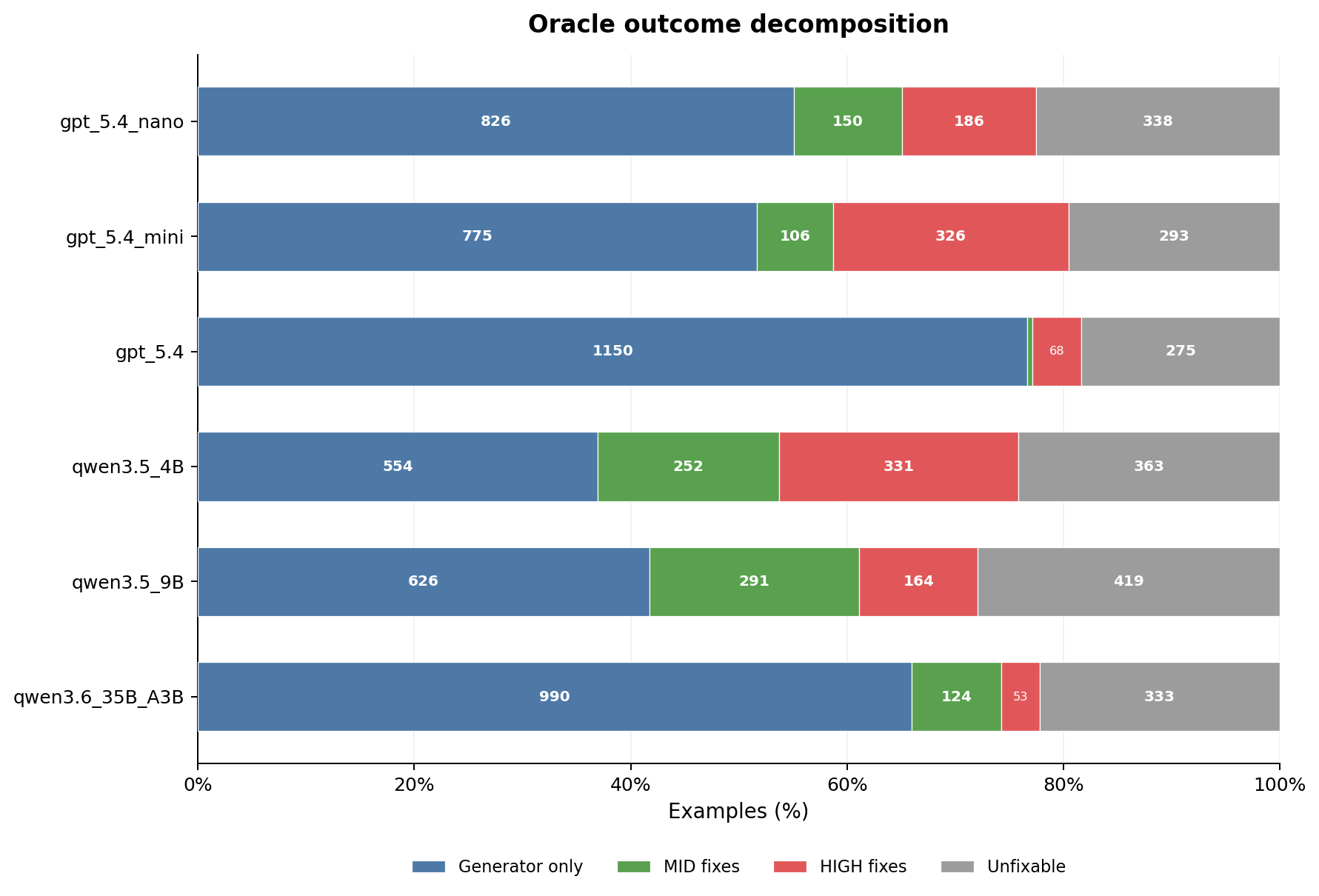}
\caption{Oracle outcome decomposition across models on a 1,500-example diagnostic sample from the FinQA training split. The bars show which examples are already correct under the generator, fixed by \textsc{Mid}, fixed only by \textsc{High}, or not fixed by any verification tier. This diagnostic decomposition is used to analyze verifier outcome structure.}
\label{fig:oracle_decomposition}
\end{figure}

Figure~\ref{fig:oracle_decomposition} complements the Pareto frontiers in Figure~\ref{fig:pareto_all_models} by decomposing each model's 1,500 randomly sampled FinQA examples (training split only) into oracle outcome categories: the blue region contains examples already answered correctly by the generator, for which additional verification is unnecessary; the green and red regions contain examples where verification is useful, with \textsc{Mid} and \textsc{High} respectively fixing generator errors; and the gray region contains examples that none of the available verifiers can fix. This decomposition explains the structure of the accuracy-cost frontiers. Models with a large blue region, such as \texttt{gpt-5.4} and \texttt{Qwen3.6-35B-A3B}, already answer many examples correctly, so the main value of BudgetVerify is avoiding unnecessary verification while selectively correcting the remaining errors. In contrast, weaker models such as \texttt{Qwen3.5-4B} have much larger green and red regions, indicating more opportunities for verification to improve accuracy. This corresponds to the larger upward movement in their Pareto frontiers.

\subsection{Main Results}

\begin{figure*}[t]
\centering

\begin{minipage}[t]{0.48\textwidth}
\centering

\begin{subfigure}[t]{1\linewidth}
    \centering
    \includegraphics[width=\linewidth]{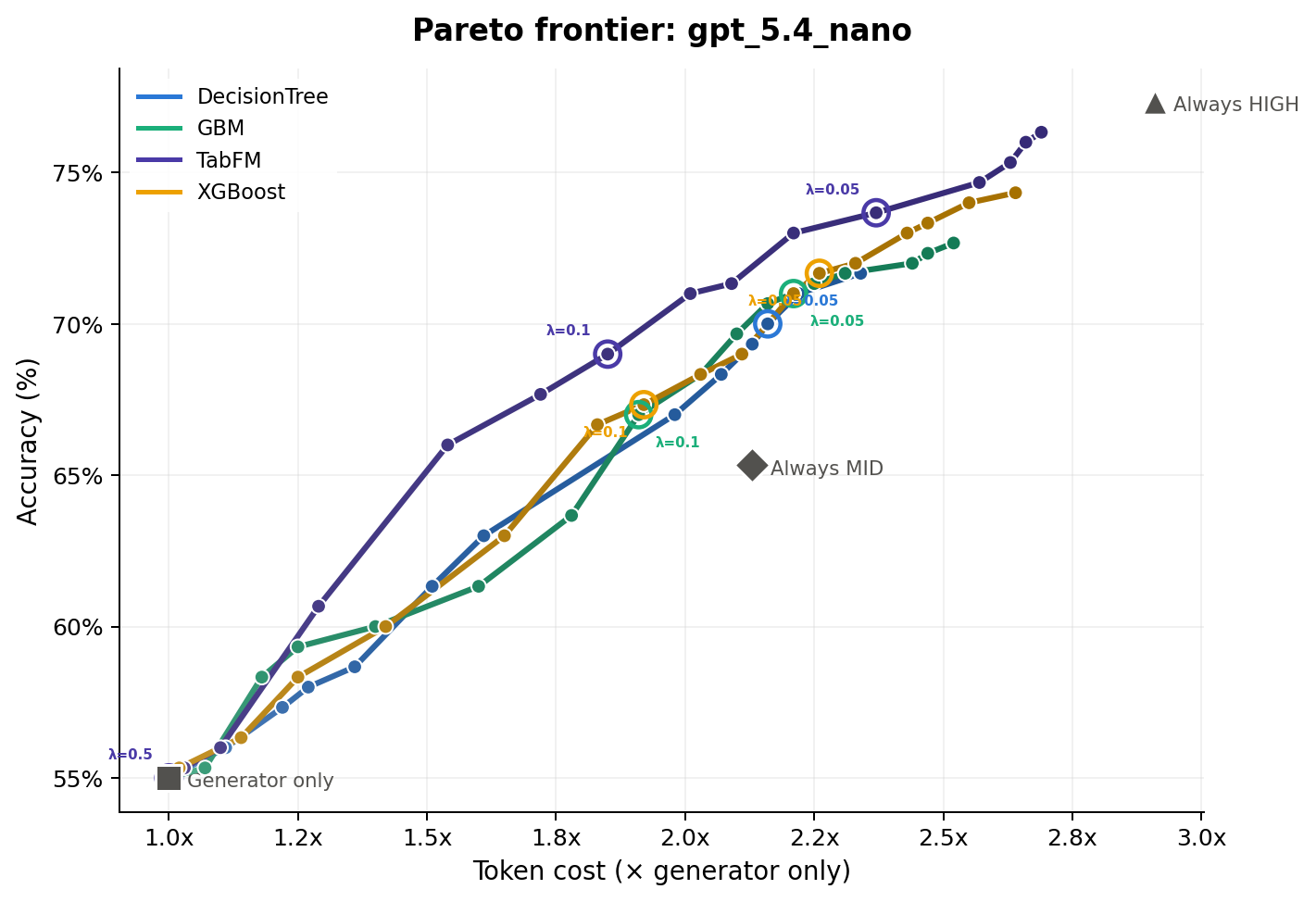}
    \caption{\texttt{gpt-5.4-nano}}
    \label{fig:pareto_gpt54_nano}
\end{subfigure}

\vspace{0.6em}

\begin{subfigure}[t]{1\linewidth}
    \centering
    \includegraphics[width=\linewidth]{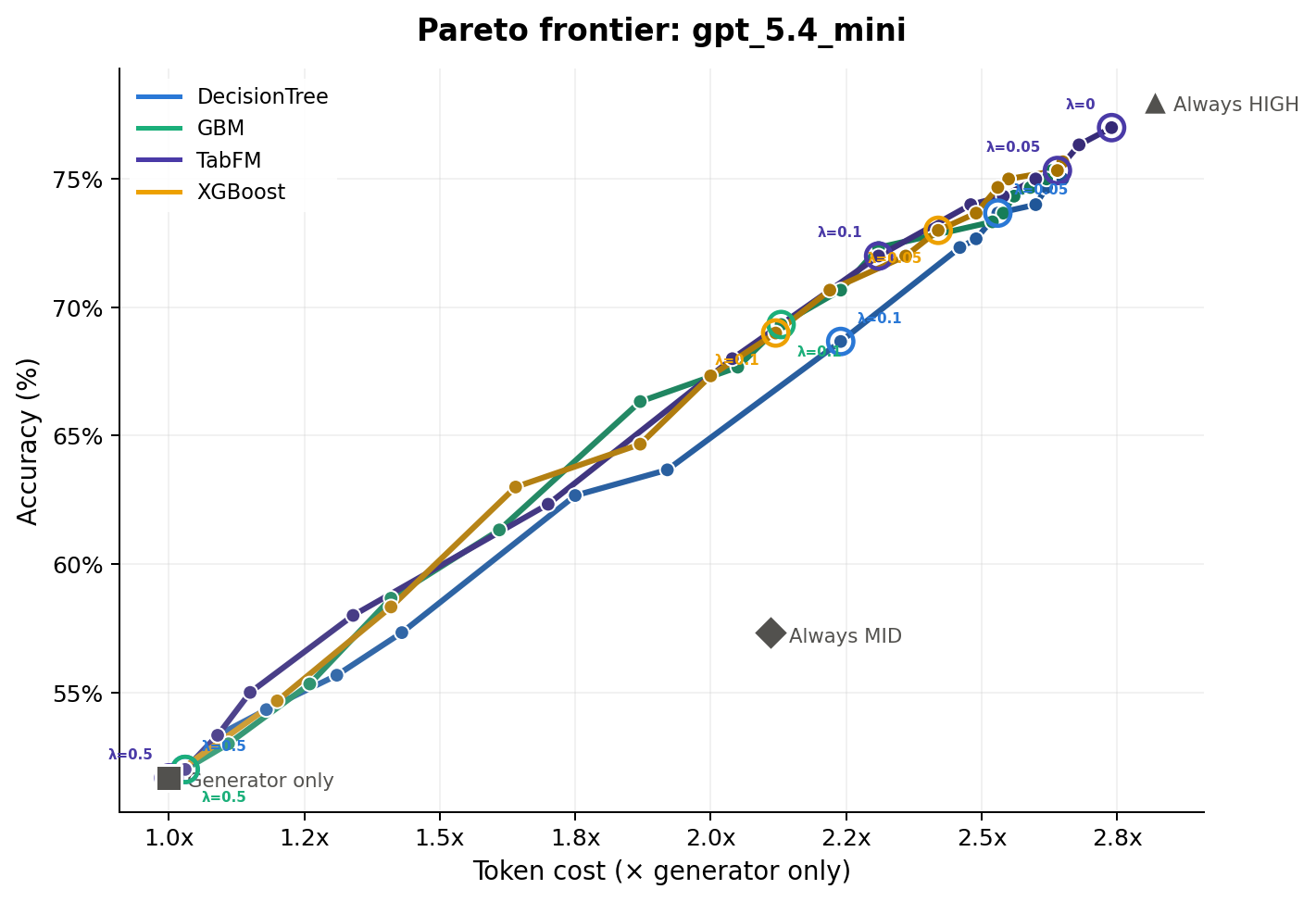}
    \caption{\texttt{gpt-5.4-mini}}
    \label{fig:pareto_gpt54_mini}
\end{subfigure}

\vspace{0.6em}

\begin{subfigure}[t]{1\linewidth}
    \centering
    \includegraphics[width=\linewidth]{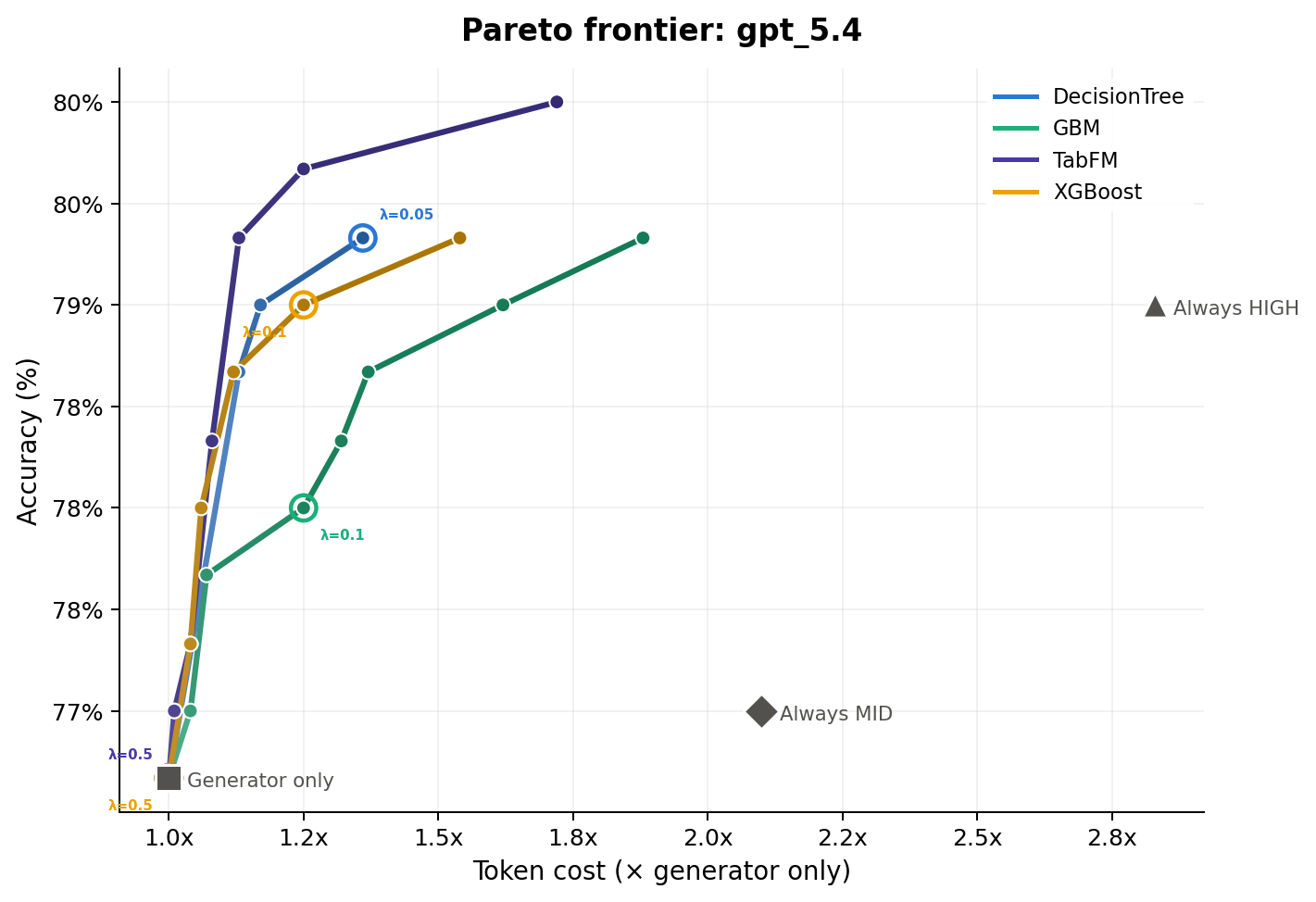}
    \caption{\texttt{gpt-5.4}}
    \label{fig:pareto_gpt54}
\end{subfigure}
\end{minipage}
\hfill
\begin{minipage}[t]{0.48\textwidth}
\centering

\begin{subfigure}[t]{1\linewidth}
    \centering
    \includegraphics[width=\linewidth]{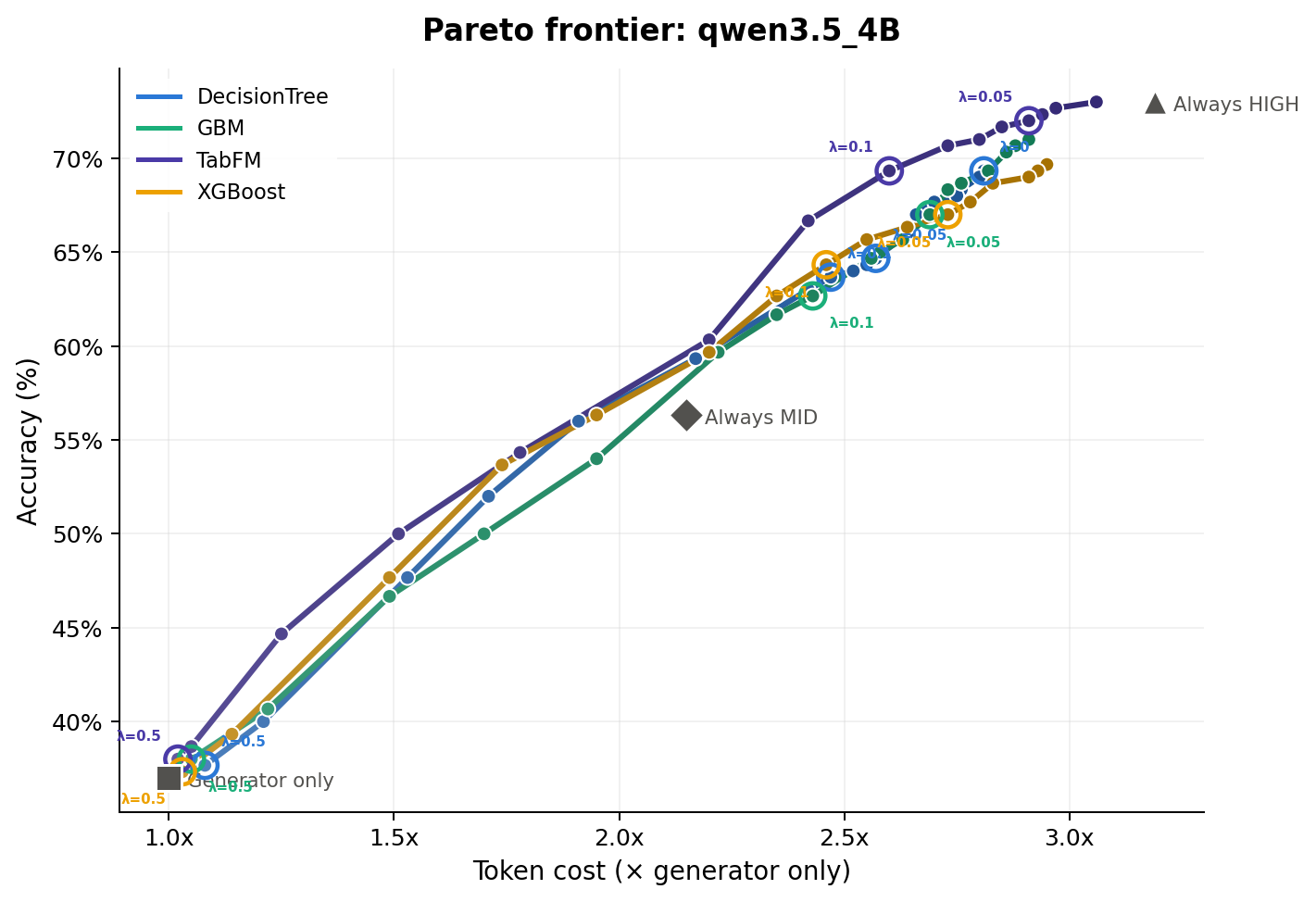}
    \caption{\texttt{Qwen3.5-4B}}
    \label{fig:pareto_qwen35_4b}
\end{subfigure}

\vspace{0.6em}

\begin{subfigure}[t]{1\linewidth}
    \centering
    \includegraphics[width=\linewidth]{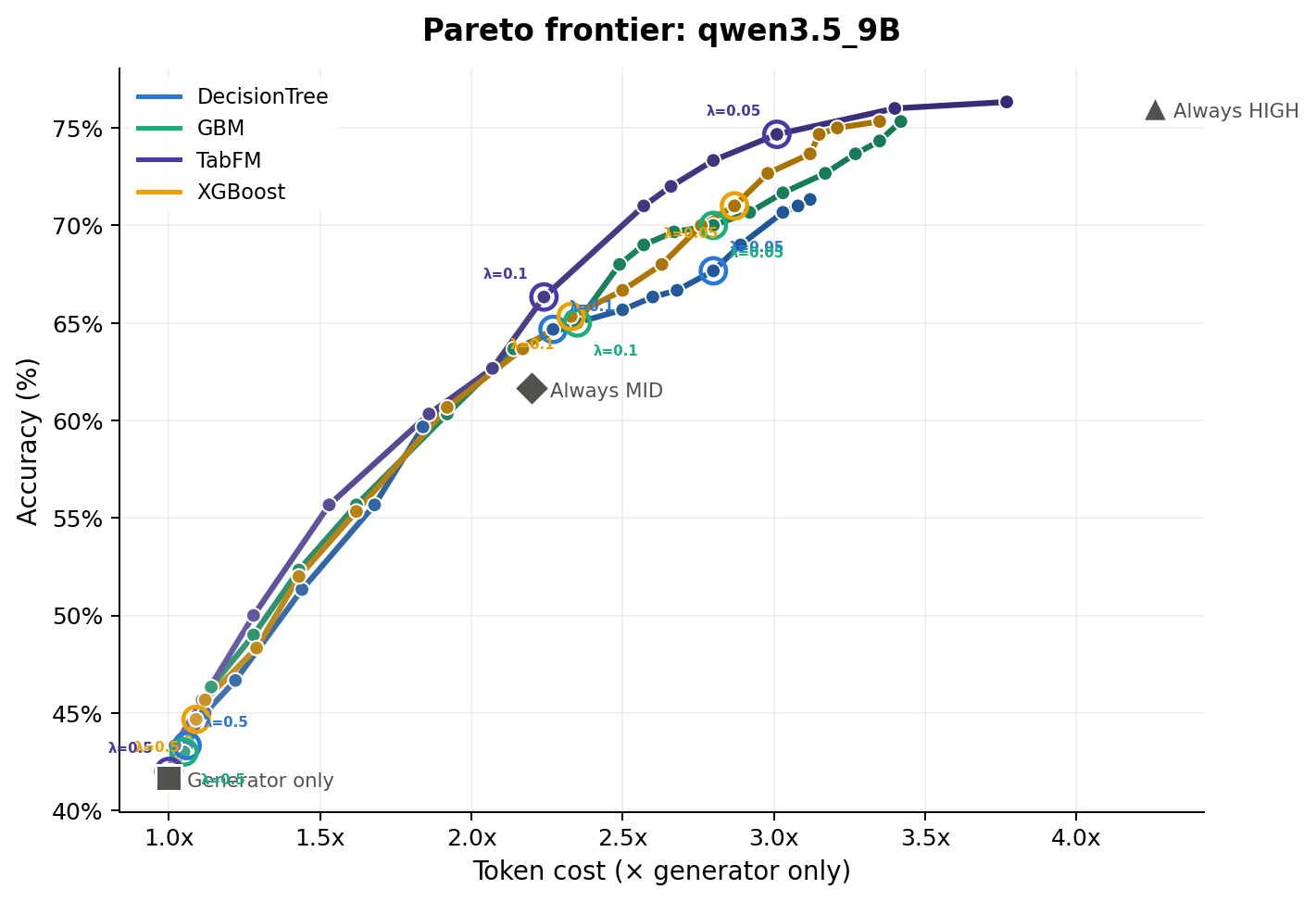}
    \caption{\texttt{Qwen3.5-9B}}
    \label{fig:pareto_qwen35_9b}
\end{subfigure}

\vspace{0.6em}

\begin{subfigure}[t]{1\linewidth}
    \centering
    \includegraphics[width=\linewidth]{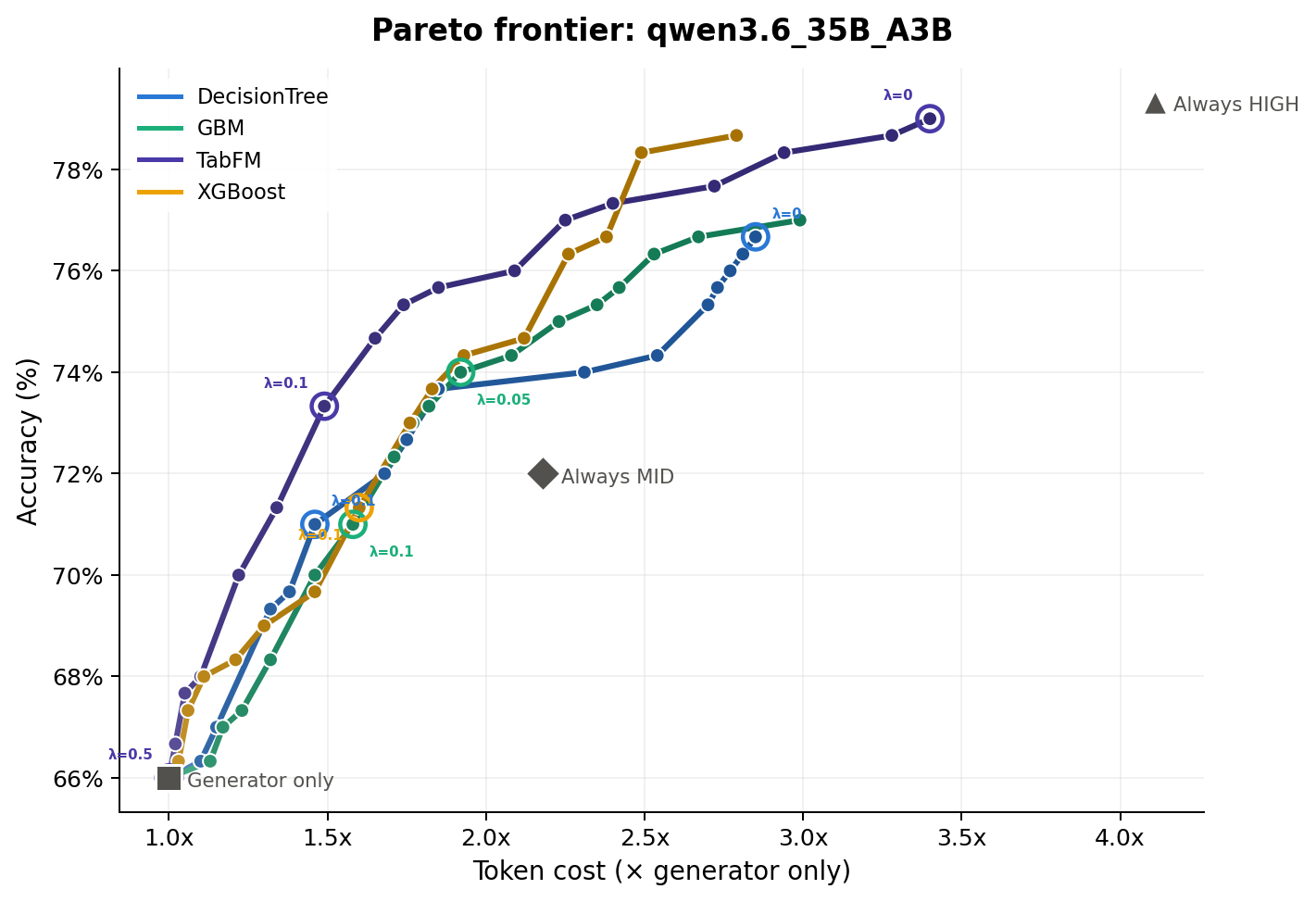}
    \caption{\texttt{Qwen3.6-35B-A3B}}
    \label{fig:pareto_qwen36_35b_a3b}
\end{subfigure}
\end{minipage}

\caption{
Within-model accuracy-token-cost tradeoff curves across six base-model settings.
Each panel should be interpreted as an independent replication of the same experiment,
rather than as a comparison of absolute model performance across panels. The x-axis
reports total token cost normalized by the generator-only \textsc{Low} baseline for that
model, and the y-axis reports final answer accuracy on the FinQA test split. Colored
curves correspond to supervised routing policies obtained by sweeping the cost parameter
$\lambda$. Each colored point represents a separately trained router $\pi_\lambda$, since
each $\lambda$ induces a different set of cost-aware oracle labels. Smaller $\lambda$ values
favor accuracy and route more examples to \textsc{Mid} or \textsc{High}, while larger
$\lambda$ values favor lower-cost \textsc{Low} decisions. Gray markers denote fixed
verification baselines: \textsc{Generator Only}, \textsc{Always Mid}, and \textsc{Always High}.
The upper envelope of the routed points gives the empirical accuracy--cost frontier.
}
\label{fig:pareto_all_models}
\end{figure*}

Figure~\ref{fig:pareto_all_models} shows the within-model accuracy-cost tradeoff across
all six base-model settings. Each colored point corresponds to one value of the cost
parameter $\lambda$ and one separately trained router $\pi_\lambda$, since each $\lambda$
induces a different set of cost-aware oracle labels. Smaller $\lambda$ values favor accuracy
and route more examples to \textsc{Mid} or \textsc{High}, while larger values favor cheaper
\textsc{Low} decisions.

Across all six base models, BudgetVerify yields a more efficient accuracy-cost frontier than
the fixed verification baseline policies. Routed policies improve over
\textsc{Generator Only}, generally outperform \textsc{Always Mid}, and can approach
\textsc{Always High} accuracy at lower normalized token cost. In some settings, they
even reach or surpass the \textsc{Always High} operating point, suggesting that uniformly
applying the strongest verifier can waste compute or revise answers that were already
correct. The smooth progression of the curves as $\lambda$ varies further shows that the
routers learn meaningful cost-sensitive allocation policies rather than collapsing to a
single verification tier.

\subsubsection{Adaptive routing improves the accuracy-cost frontier across base models}



The absolute position of the frontier depends on the base model, but the qualitative pattern
is consistent. Stronger generators, such as \texttt{gpt-5.4} and \texttt{Qwen3.6-35B-A3B},
start from higher generator-only accuracy. For these models, the main benefit of routing is
to preserve already-correct answers and avoid unnecessary calls to \textsc{High} verifier, while still
sending selected difficult cases to stronger verification. Weaker generators, such as
\texttt{gpt-5.4-nano}, \texttt{gpt-5.4-mini}, \texttt{Qwen3.5-4B}, and \texttt{Qwen3.5-9B},
start from lower generator-only accuracy and therefore offer more opportunities for
\textsc{Mid} or \textsc{High} verifier to repair errors.

Despite these differences, the same conclusion holds across both commercial and
open-weight models: adaptive verification provides better accuracy at a given token budget
than applying a fixed verification tier uniformly. This supports the central claim that the
value of verification is input-, model-, and answer-dependent rather than constant across samples.

\subsubsection{Comparison across router classifiers}

We evaluate four supervised routers for selecting among the \textsc{Low}, \textsc{Mid}, and
\textsc{High} verification tiers. Decision Tree provides a simple interpretable baseline that
learns threshold-style routing rules from the extracted features. GBM and XGBoost are
gradient-boosted tree methods that capture nonlinear feature interactions while remaining
lightweight, making them strong tabular-learning baselines for cost-aware routing. TabFM
\citep{tabfm2026} is included to test whether a more expressive pretrained tabular model
can better learn the relationship between question features, generator metadata, and the
expected value of additional verification.

Among the supervised routers, TabFM produces the strongest and most consistent frontiers.
It is frequently the upper envelope across the GPT and Qwen settings, especially for
\texttt{gpt-5.4-nano}, \texttt{Qwen3.5-4B}, \texttt{Qwen3.5-9B}, and
\texttt{Qwen3.6-35B-A3B}. XGBoost is generally the next most competitive method, often
tracking TabFM closely and achieving strong mid-budget tradeoffs. GBM is also competitive
but tends to fall slightly below TabFM and XGBoost in the high-accuracy region. Decision
Tree is the weakest router overall, particularly in mid- and high-budget regimes, although
it remains a useful interpretable baseline. This gap suggests that verification routing
depends on nonlinear interactions among question cues, generator metadata, predicted
operation type, evidence coverage, and cost, rather than on a small number of simple
threshold rules.


\section{Conclusions}
\label{sec:conclusions}
We introduced BudgetVerify, a budget-tiered verification framework for financial QA. Rather than applying the same verifier to every generated answer, BudgetVerify learns to select among verification tiers: no verification, lightweight check-and-revise, and solve-first verification using only pre-verification features. Experiments on FinQA across six commercial and open-weight base models show that adaptive verification yields more efficient Pareto frontiers than fixed verification: BudgetVerify consistently improves upon uniform \textsc{Mid} verification and often matches or approaches uniform \textsc{High} verification while using fewer tokens. 

The purpose of evaluating multiple base models is not to compare or rank their absolute capabilities, but to test whether the same trade-off is robust across model settings. Across every evaluated base model, we observe the same qualitative Pareto-frontier shape: selective verification produces substantial initial accuracy gains, followed by diminishing returns as increasingly expensive verification is applied. This pattern reflects the input- and answer-dependent value of verification: easy or already-correct answers often benefit from being left unchanged, while more difficult numerical cases are more likely to benefit from stronger verification. The repeated frontier geometry suggests that budget-aware verification is a general inference-time allocation principle rather than a property of a particular model family.



\bibliography{colm2026_conference}
\bibliographystyle{colm2026_conference}

\clearpage
\appendix
\section{Appendix}

{\small
\renewcommand{\_}{\textunderscore\allowbreak}%
\begin{xltabular}{\linewidth}{p{0.18\linewidth} p{0.27\linewidth} X}
\caption{Pre-verification feature dictionary for the BudgetVerify router. All features are derived from the question, lightweight context statistics, or the generator's structured answer before executing \textsc{Mid} or \textsc{High}. Outcome fields such as verifier correctness, oracle labels, and post-verification outputs are excluded from router inputs.}
\label{tab:router_features} \\

\toprule
\textbf{Category} & \textbf{Feature} & \textbf{Definition} \\
\midrule
\endfirsthead
\multicolumn{3}{l}{\tablename~\thetable\ (continued)}\\
\toprule
\textbf{Category} & \textbf{Feature} & \textbf{Definition} \\
\midrule
\endhead
\midrule
\multicolumn{3}{r}{\footnotesize continued on next page}\\
\endfoot
\bottomrule
\addlinespace[2ex]
\endlastfoot

\multicolumn{3}{l}{\textbf{Question features}} \\
\midrule
Question & \texttt{q\_word\_count} & Number of words in the question. \\
Question & \texttt{q\_char\_count} & Number of characters in the question. \\
Question & \texttt{q\_number\_count} & Number of numeric strings in the question, such as years, amounts, or percentages. \\
Question & \texttt{q\_period\_count} & Number of distinct years, fiscal years, quarters, or periods mentioned in the question. \\
Question & \texttt{q\_has\_percent} & Indicator for whether the question contains \texttt{\%}, \texttt{percent}, or \texttt{percentage}. \\
Question & \texttt{q\_has\_currency} & Indicator for whether the question contains monetary cues such as \texttt{\$}, \texttt{dollar}, or \texttt{USD}. \\
Question & \texttt{q\_has\_calc\_terms} & Indicator for calculation cues such as \texttt{increase}, \texttt{decrease}, \texttt{growth}, \texttt{difference}, \texttt{change}, \texttt{ratio}, \texttt{margin}, \texttt{sum}, or \texttt{average}. \\
Question & \texttt{q\_has\_comparison\_terms} & Indicator for comparison cues such as \texttt{higher}, \texttt{lower}, \texttt{greater}, \texttt{less}, \texttt{largest}, \texttt{smallest}, \texttt{compare}, or \texttt{versus}. \\
Question & \texttt{q\_has\_financial\_terms} & Indicator for financial terms such as \texttt{revenue}, \texttt{income}, \texttt{expense}, \texttt{assets}, \texttt{cash}, \texttt{debt}, \texttt{equity}, \texttt{profit}, or \texttt{loss}. \\
Question & \texttt{q\_has\_ratio\_terms} & Indicator for denominator-sensitive cues such as \texttt{ratio}, \texttt{portion}, \texttt{share}, \texttt{of total}, or \texttt{as a percentage of}. \\
Question & \texttt{q\_has\_margin\_terms} & Indicator for margin-related cues such as \texttt{margin}, \texttt{gross margin}, or \texttt{operating margin}. \\
Question & \texttt{q\_has\_lookup\_terms} & Indicator for direct extraction wording such as \texttt{what was}, \texttt{how much was}, or \texttt{amount of}. \\

\midrule
\multicolumn{3}{l}{\textbf{Context-light features}} \\
\midrule
Context & \texttt{context\_word\_count} & Number of words in the serialized context. The implementation builds the context from \texttt{pre\_text}, a pipe-table rendering of the table, and \texttt{post\_text}. \\
Context & \texttt{context\_char\_count} & Number of characters in the serialized context. \\
Context & \texttt{context\_number\_count} & Number of numeric strings in the serialized context. \\
Context & \texttt{context\_period\_count} & Number of distinct years, quarters, or periods in the serialized context. \\
Context & \texttt{context\_table\_row\_count} & Number of rows in the FinQA table. \\
Context & \texttt{context\_table\_col\_count} & Number of columns in the FinQA table. \\
Context & \texttt{context\_number\_density} & Ratio of numeric strings to total words in the context, i.e., \texttt{context\_number\_count / max(context\_word\_count, 1)}. \\
Context & \texttt{context\_estimated\_tokens} & Estimated token count of the serialized context; useful for cost-aware routing. \\

\midrule
\multicolumn{3}{l}{\textbf{Generator answer features}} \\
\midrule
Generator output & \texttt{answer\_is\_numeric} & Indicator for whether the generator's \texttt{final\_answer} contains a numeric value or \texttt{answer\_type = numeric}. \\
Generator output & \texttt{answer\_is\_yes\_no} & Indicator for whether the generated answer is \texttt{yes} or \texttt{no}. \\
Generator output & \texttt{answer\_number\_count} & Number of numeric strings in the generator's \texttt{final\_answer}. \\
Generator output & \texttt{answer\_unit\_number} & Indicator for \texttt{answer\_unit = number}. \\
Generator output & \texttt{answer\_unit\_percent} & Indicator for \texttt{answer\_unit = percent}. \\
Generator output & \texttt{answer\_unit\_dollars} & Indicator for \texttt{answer\_unit = dollars}. \\
Generator output & \texttt{answer\_unit\_ratio} & Indicator for \texttt{answer\_unit = ratio}. \\
Generator output & \texttt{answer\_unit\_unknown} & Indicator for \texttt{answer\_unit = unknown}. \\
Generator output & \texttt{confidence\_high} & Indicator for generator confidence \texttt{high}. \\
Generator output & \texttt{confidence\_medium} & Indicator for generator confidence \texttt{medium}. \\
Generator output & \texttt{confidence\_low} & Indicator for generator confidence \texttt{low}. \\

\midrule
\multicolumn{3}{l}{\textbf{Cross-risk features}} \\
\midrule
Cross-risk & \texttt{multi\_period\_numeric} & Indicator for \texttt{q\_period\_count >= 2} and \texttt{answer\_is\_numeric = 1}. \\
Cross-risk & \texttt{derived\_metric\_signal} & Indicator for whether the question contains calculation cues or the generator provides a formula. \\
Cross-risk & \texttt{calc\_question\_no\_formula} & Indicator for calculation-like questions where the generator did not provide a formula. \\
Cross-risk & \texttt{numeric\_answer\_unknown\_unit} & Indicator for numeric answers with \texttt{answer\_unit = unknown}. \\
Cross-risk & \texttt{percent\_question\_nonpercent\_answer} & Indicator for percentage questions where the generator's answer unit is not percent. \\
Cross-risk & \texttt{ratio\_question\_percent\_answer} & Indicator for ratio questions where the generator answers as a percentage. \\
Cross-risk & \texttt{period\_mismatch\_signal} & Indicator for questions mentioning periods when the generator lists no periods or mismatched periods. \\
Cross-risk & \texttt{formula\_without\_operands} & Indicator for formula present but no operands listed. \\
Cross-risk & \texttt{operands\_without\_sources} & Indicator for operands listed without source evidence. \\
Cross-risk & \texttt{portion\_ratio\_signal} & Indicator for denominator-sensitive wording such as \texttt{portion}, \texttt{share}, \texttt{of total}, or \texttt{as a percentage of}. \\
Cross-risk & \texttt{signed\_change\_signal} & Indicator for signed-change wording such as \texttt{decline}, \texttt{decrease}, \texttt{fell}, or \texttt{change}. \\

\end{xltabular}
}

\clearpage

\begin{center}
\begin{minipage}{\textwidth}
\begin{verifierbox}
MID verifier: candidate-centered check-and-revise

Input:
  - Question
  - Full context
  - Generator answer

Procedure:
  1. Treat the generator answer as the default final answer.
  2. Check only for obvious local errors:
       - malformed final answer,
       - copied wrong value,
       - copied wrong period,
       - simple arithmetic inconsistency,
       - obvious yes/no mismatch.
  3. Use only the candidate's stated operands, formula, or evidence when possible.
  4. If the answer is plausible, keep it.
  5. Revise only when the correction is clear and directly supported.

Output:
  - KEEP or REVISE
  - final answer
  - brief error type
  - confidence
\end{verifierbox}

\captionof{figure}{\textsc{Mid} verifier template. The verifier performs lightweight candidate-centered check-and-revise.}
\label{fig:mid_template}
\end{minipage}
\end{center}

\begin{center}
\begin{minipage}{\textwidth}
\begin{verifierbox}
HIGH verifier: problem-centered solve-and-compare

Input:
  - Question
  - Full context
  - Generator answer, used only after independent solving

Stage 1: Independent solve
  1. Hide the generator answer.
  2. Solve the problem from the context.
  3. Extract the relevant values, periods, operation, formula, and answer unit.
  4. Return a structured solution with operation, operands, formula, and confidence.

Stage 2: Deterministic computation
  1. Execute the extracted operation with a calculator.
  2. Support operations such as lookup, difference, percent change, ratio,
     margin, sum, average, and comparison.
  3. Format the computed answer.

Stage 3: Grounding check
  1. Check whether extracted operand values appear in the context.
  2. Lower confidence if operands are ungrounded.
  3. Fall back conservatively if the independent solution has no usable signal.

Stage 4: Candidate comparison
  1. Compare the generator answer with the independently computed answer.
  2. KEEP if they match after valid rounding or formatting.
  3. REVISE if they differ and the independent solution is grounded.

Output:
  - KEEP or REVISE
  - final answer
  - candidate match indicator
  - brief error type
  - confidence
\end{verifierbox}

\captionof{figure}{\textsc{High} verifier template. The verifier solves the question independently from the context, executes the extracted calculation with a deterministic calculator, checks operand grounding, and then compares the computed answer with the generator's candidate to decide whether to keep or revise it.}
\label{fig:high_template}
\end{minipage}
\end{center}

\end{document}